\documentclass[sigconf]{acmart}
\usepackage{enumitem}
\usepackage{rotating,amsmath,amsfonts,amsthm}
\usepackage{epsfig,mathrsfs,natbib,color,graphics}
\usepackage{graphicx}
\usepackage{array,wrapfig}
\usepackage{hyperref}
\usepackage{booktabs}
\usepackage{multirow}
\usepackage{subcaption}

\AtBeginDocument{%
  \providecommand\BibTeX{{%
    \normalfont B\kern-0.5em{\scshape i\kern-0.25em b}\kern-0.8em\TeX}}}
    
\begin{document}
\title{Self-supervised learning for fast and scalable time series hyper-parameter tuning}


\author{Peiyi Zhang}
\authornote{Both authors contributed equally to this research.}
\affiliation{%
  \institution{Purdue University}
  \city{West Lafayette}
  \state{Indiana}
  \country{USA}
  \postcode{47907}
}
\email{zhan2763@purdue.edu}

\author{Xiaodong Jiang}
\authornotemark[1]
\affiliation{%
  \institution{Facebook Inc.}
  \streetaddress{1 Hacker Way}
  \city{Menlo Park}
  \country{USA}}
\email{iamxiaodong@fb.com}

\author{Ginger M Holt}
\affiliation{%
  \institution{Facebook Inc.}
  \streetaddress{1 Hacker Way}
  \city{Menlo Park}
  \country{USA}}
\email{gingermholt@fb.com}

\author{Nikolay Pavlovich Laptev}
\affiliation{%
  \institution{Facebook Inc.}
  \streetaddress{1 Hacker Way}
  \city{Menlo Park}
  \country{USA}}
\email{nlaptev@fb.com}

\author{Caner Komurlu}
\affiliation{%
  \institution{Facebook Inc.}
  \streetaddress{1 Hacker Way}
  \city{Menlo Park}
  \country{USA}}
\email{ckomurlu@fb.com}

\author{Peng Gao}
\affiliation{%
  \institution{Facebook Inc.}
  \streetaddress{1 Hacker Way}
  \city{Menlo Park}
  \country{USA}}
\email{penggao@fb.com}

\author{Yang Yu}
\affiliation{%
  \institution{Facebook Inc.}
  \streetaddress{1 Hacker Way}
  \city{Menlo Park}
  \country{USA}}
\email{yangbk@fb.com}



\begin{abstract}
Hyper-parameters of time series models play an important role in time series analysis. Slight differences in hyper-parameters might lead to very different forecast results for a given model, and therefore, selecting good hyper-parameter values is indispensable. Most of the existing generic hyper-parameter tuning methods, such as Grid Search, Random Search, Bayesian Optimal Search, are based on one key component - search, and thus they are computationally expensive and cannot be applied to fast and scalable time-series hyper-parameter tuning (HPT). We propose a self-supervised learning framework for HPT (SSL-HPT), which uses time series features as inputs and produces optimal hyper-parameters. SSL-HPT algorithm is 6-20x faster at getting hyper-parameters compared to other search based algorithms  while producing comparable accurate forecasting results in various applications.
\end{abstract}




\maketitle
\section{Introduction}

Forecasting is a data science and machine learning task which plays a role in many practical domains, from capacity planning and management \cite{huang2008demand,taylor2018forecasting}, demand forecasting \cite{laptev2017time}, energy prediction \cite{grolinger2016energy}, to anomaly detection \cite{malhotra2016lstm,laptev2017time}. The rapid advances in computing technologies have enabled businesses to keep track of large number of time series datasets. Hence, the need to regularly forecast millions of time series is becoming increasingly high. It remains challenging to obtain fast and accurate forecasts for a large number of time series data sets. Inspired by meta-learning techniques in \cite{talagala2018meta} and recent advancements of self-supervised learning \cite{lan2019albert,jing2020self,zhai2019s4l}, we propose a \textbf{self-supervised learning (SSL)} framework, which provides accurate forecasts with lower computational time and resources. This SSL framework is designed for both model selection and hyper-parameter tuning, with details explanations in Section \ref{sec:ms} and Section \ref{sec:hpt}, respectively.

\subsection{Model selection}
\label{sec:ms}
We summarize the existing approaches of model selection for a large collection of time series datasets as follows: 
\begin{itemize}[noitemsep,topsep=0pt]
 \item[1. ]Pre-selected model: researchers and scientists may choose a specific model that works well for certain datasets, based on their domain knowledge.
 \item[2. ]Randomly chosen model: without extensive parameter tuning, we might also randomly select a forecasting model without good understanding of the model or datasets.
 \item[3. ]Ensemble model: leverage all candidate models and perform ensembling. Our proposed self-supervised learning framework is different from approaches above, given a time series data, we train a classifier to pick the best performing model. 
\end{itemize}

The self-supervised learning framework for model selection (SSL-MS) consists of three steps: 
\begin{itemize}[noitemsep,topsep=0pt]
 \item[1. ]Offline training data preparation. We obtain (a). time series features for each time series, and (b). the best performing model for each time series via offline exhaustive hyper-parameter tuning.
 \item[2. ]Offline training. A classifier (self-supervised learner) is trained with the data from Step (1), where the input feature (predictor) is the time series feature, the label is the best performing model. 
 \item[3. ]Online model prediction. In our online services, for a new time series data, we first extract features, then make inference with our pre-trained classifier, such as random forest.
 \end{itemize}

\subsection{Hyper-parameter tuning}
\label{sec:hpt}
In practice, model selection is not enough, we usually need to tune the model for a specific data to obtain good performance. In time series analysis, in order to achieve a good forecasting accuracy, it’s important to set “good” hyper-parameters for a given time series model. Typically, values of hyper-parameters are often specified by practitioners after a tuning procedure, such as Random search, Grid search, or Bayesian Optimal Search. Grid search searches the optimal hyper-parameter over all possible hyper-parameter combinations from defined hyper-parameter value space, while random search performs a search by random steps. However, for some models, the number of possible hyper-parameter combinations is extremely large, and thus it’s not feasible to use grid search. Another major disadvantage is that both grid and random search pay no attention to past results at all and would keep searching across the entire space even though it’s clear the “optimal” one lies in a small region! In contrast to grid and random search, Bayesian Optimal Search (BOS) is able to keep track of past evaluation results and choose the next hyper-parameters in an informed manner, and thus be more efficient. Given an initial point, BOS iteratively searches for next best point until max iterations is reached. 

Although the three existing search strategies are useful and perform well for a single time series dataset, they are computationally expensive thus cannot achieve fast and scalable forecasting tasks with large scale time-series datasets. Hence, in real word applications, instead of using exhaustive tuning, randomly chosen hyper-parameters (random-HP) is commonly used. 

Alternative to exhaustive tuning and random-HP, we propose a self-supervised learning framework for HPT (SSL-HPT). It uses time series features as inputs and the most promising hyper-parameters, for a given model, as outputs. The SSL-HPT framework consists of three steps: (1). Offline training data preparation. Similar to SSL-MS, we also need to obtain the time series features, then perform offline exhaustive parameter tuning to get the best performed hyper-parameters for each model and data combination; (2). Offline training. A multi-task neural network (self-supervised learner) is trained with the datasets from Step (1) for each model; (3). Online hyper-parameters tuning. In our online system, for a new time series data, we first extract features, then make inference with our pre-trained multi-task neural network. Since SSL-HPT takes constant time to choose hyper-parameters, it makes fast and accurate forecasts at large scale become feasible. 

The remaining part of this paper is organized as follows. Section \ref{sec:msmain} describes the self-supervised learning framework on forecasting model selection. Section \ref{sec:sslmain} and \ref{sec:apps} describe the self-supervised learning framework on hyper-parameter tuning and its applications. Section \ref{sec:expresults} presents the experimental results on two data sets: Facebook Infrastructure dataset and M3-competitation dataset \cite{makridakis2000m3}. Section \ref{sec:conclusion} concludes the paper with a brief discussion.

\section{Self-supervised learning on forecasting model selection (SSL-MS)}
\label{sec:msmain}
\subsection{Workflow of SSL-MS}
\begin{figure*}
  \includegraphics[width=0.9\textwidth]{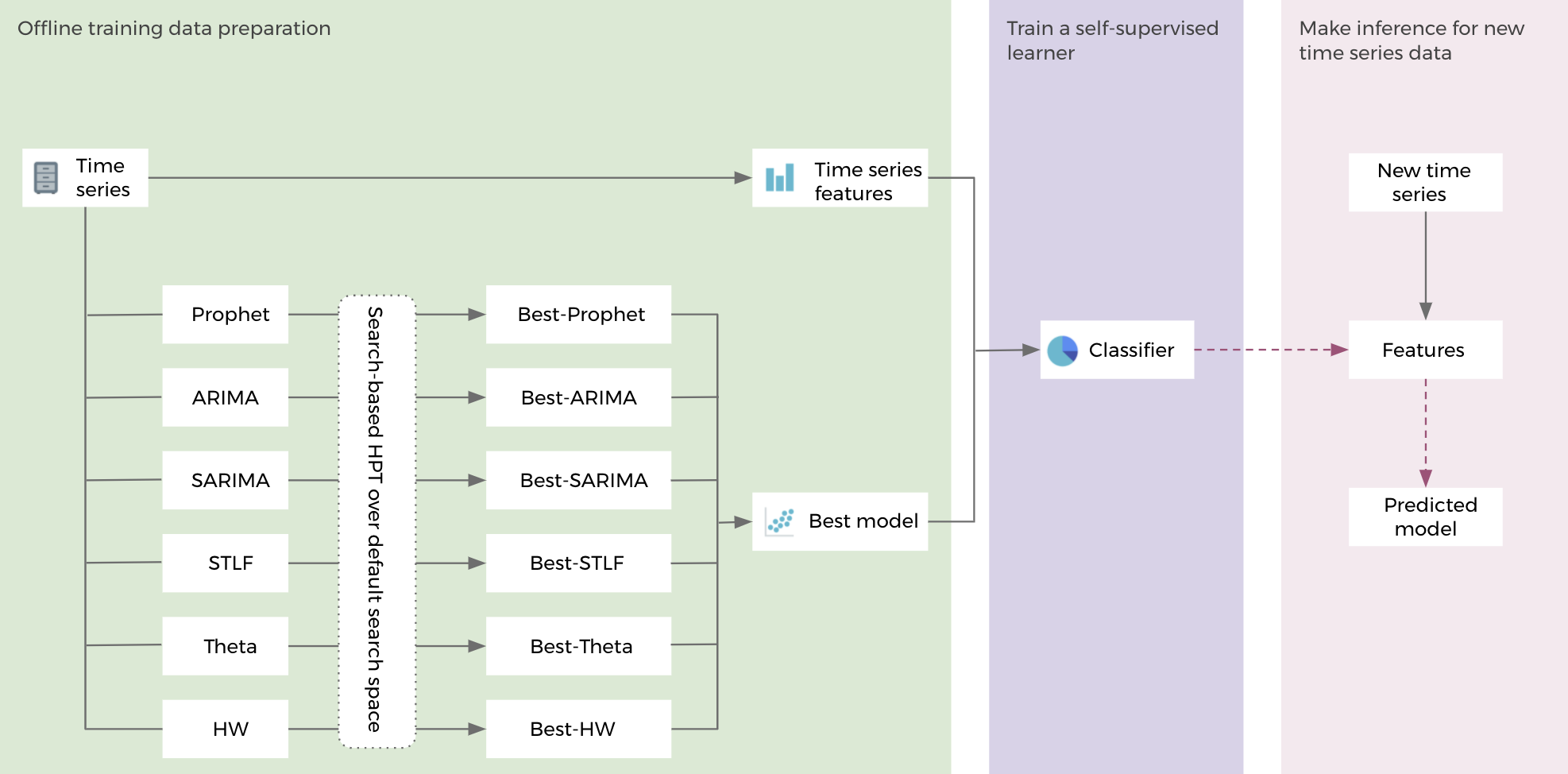}
  \caption{Workflow of Self-Supervised Learning for Model Selection (SSL-MS)} \label{fig:ssl-ms}
\end{figure*}

Figure~\ref{fig:ssl-ms} shows the workflow of SSL-MS. For each time series data, we train a set of candidate models, and identify a best parameter setting for each candidate model by hyper-parameter tuning procedures. Using these best parameter settings, we compare forecast errors over test periods from all candidate models, and then identify the best model for each time series data. The models deemed best form the output vector $Y$, and time series’ feature vectors, such as ACF/PACF based features, seasonality, linearity, Hurst exponent, ARCH statistic, form the the input vector $X$ of the classification algorithm. Once the classifier is trained, making forecast for new time series data is straightforward: calculating features and passing to the trained classifier. The choice of classification algorithm is flexible, one can choose logistic regression, random forest, GBDT, SVM, kNN, etc. 

\subsection{Time series features}
Our proposed SSL-MS algorithm requires features that enable identification of a suitable forecast model for a given time series data. Therefore, the features used should capture the dynamic structure of the time series, such as trend, peak, seasonality, autocorrelation, nonlinearity, heterogeneity, and so on. We used 40 features in our current framework and experiment, a comprehensive description of those features is presented in Appendix \ref{appendixA}.

\section{Self-supervised learning on hyper-parameter tuning (SSL-HPT)}
\label{sec:sslmain}

\subsection{Workflow of SSL-HPT}
\begin{figure*}
     \centering
     \includegraphics[width=0.9\textwidth]{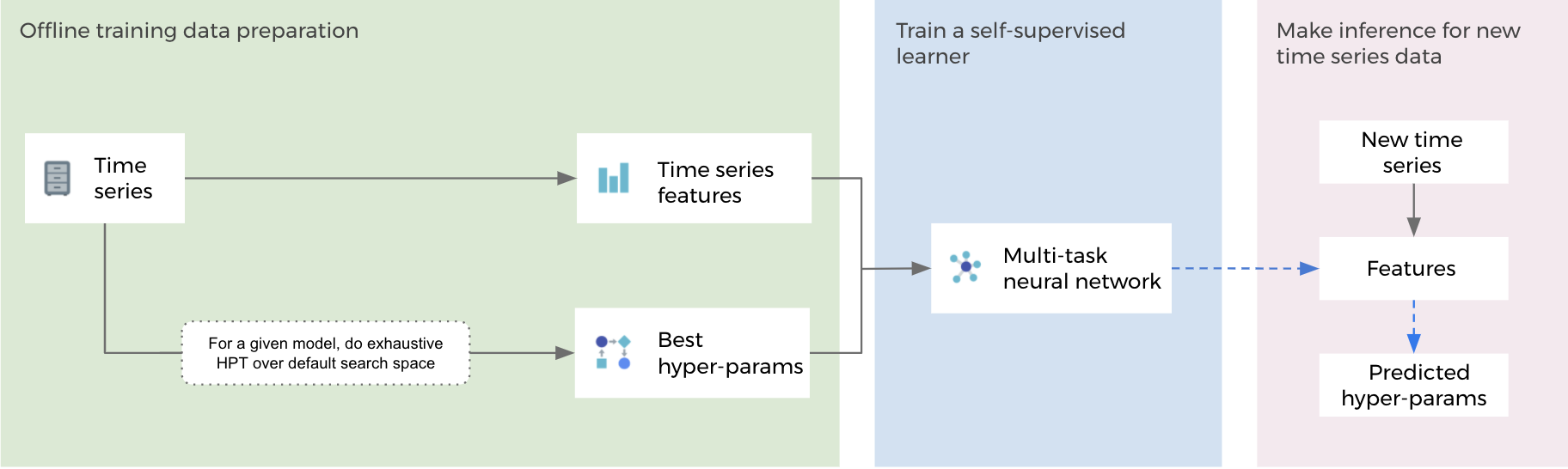}
      \caption{Workflow of Self-Supervised Learning for Hyper-Parameter Tuning (SSL-HPT)}
      \label{fig:ssl-hpt}
\end{figure*}

The workflow of SSL-MS can be naturally extended to SSL-HPT. Figure \ref{fig:ssl-hpt} shows the workflow of SSL-HPT. As this figure shows, for each time series data, given a model, all hyper-parameter settings are attempted and then the most promising one will be selected as the output $Y$. For input $X$, the time series features we used here are the same as in SSL-MS. Once the self-supervised learner is trained, for a new time series data, we can directly predict hyper-parameters and produce the forecasting results. 

\subsection{Training algorithm: multi-task neural networks}

We used the best performing model as label in the classification algorithm in SSL-MS, while the label in the SSL-HPT framework is the best performing set of hyper-parameters, for each model. Specifically, they are different in two aspects: (i) we usually have multiple hyper-parameters for a given forecasting model, so the output would be multivariate; (ii) hyper-parameters could be a combination of both categorical and numerical variables. For example, STLF model has two hyper-parameters, one is categorical while the other is numerical. For this type of label, a multi-task neural network would be the most appropriate self-supervised learner. 

We use a hard parameter sharing neural network for the multi-task learning (MTL) model \cite{ruder2017overview}. Hard parameter sharing is the most commonly used approach for MTL. It is generally applied by sharing the hidden layers between all tasks, while keeping several task-specific output layers. Using hard parameter sharing greatly reduces the risk of overfitting. In fact, the more tasks we are learning simultaneously, the more our model has to find a representation that captures all of the tasks and the less is our chance of overfitting on our original task. 

\begin{figure}
     \centering
     \includegraphics[width=0.4\textwidth]{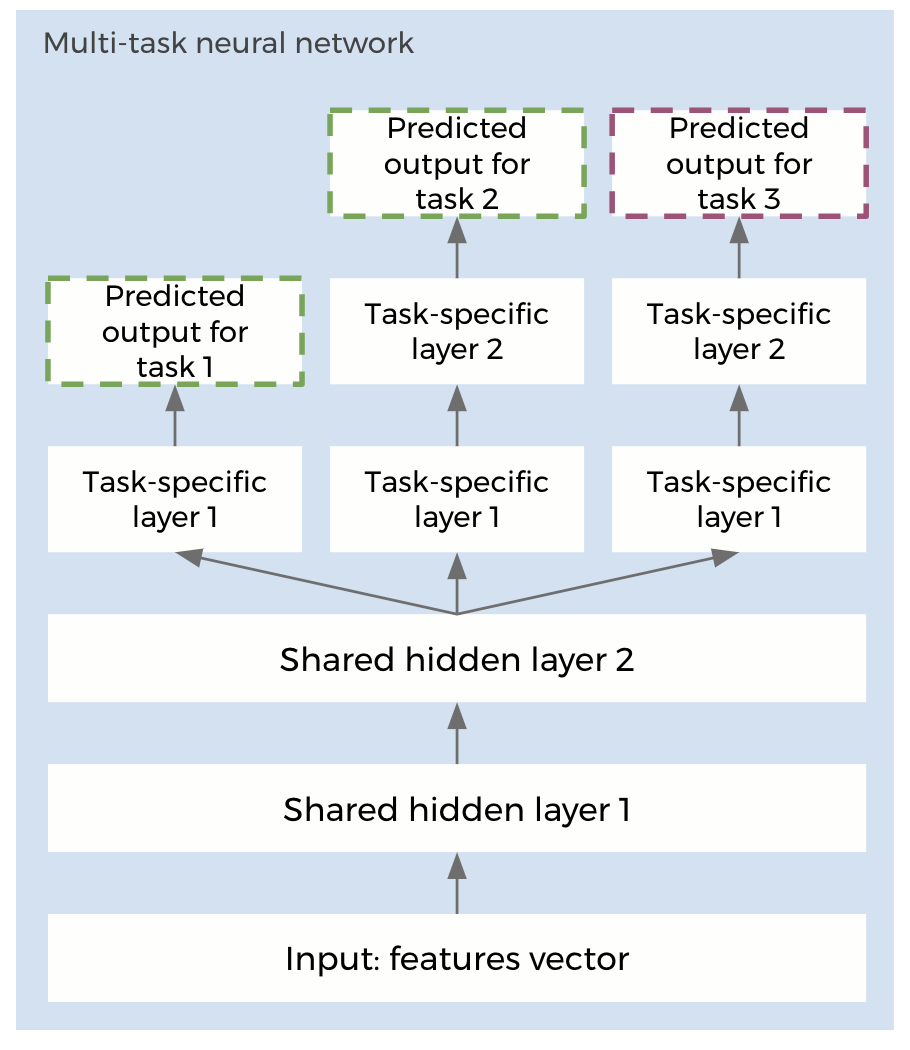}
      \caption{Illustration of a sample multi-task neural network}
      \label{fig:mnn}
\end{figure}

Suppose the output has two categorical variables, $Y_1$, $Y_2$, and one numerical variables $Y_3$, a multi-task neural network can be built as Figure \ref{fig:mnn}. In Figure \ref{fig:mnn}, two green tasks are classification tasks, while the purple one is regression task. The number of task-specific layers for each task could be different. The reason why we cannot combine two categorical $Y$s to a single categorical one is that these two might not be independent from each other. Losses of all tasks are calculated by comparing predicted values with true values. Since we use cross-entropy as the classification loss function and MSE as the regression loss function, the total loss is defined as:

\begin{equation}
Total\ Loss = Loss1 + Loss2 + Loss3/s,
\end{equation}
where $s$ is used for balancing these two different types of losses, we need to tune this parameter when we fit the neural network model. Also, $s$ helps prevent overfitting for both classification and regression tasks. After calculating the total loss, back-propagation will be applied to train this neural network. Please note, there are different ways to combine losses from each task to an overall loss, we choose such weighted summation due to its simplicity and interpretability.

\subsection{Generalization}
The SSL-HPT algorithm turns out to be a generalization of two most common HPT approaches: exhaustive tuning and random-HP. Suppose that we have $N$ time series datasets, and it takes $c$, $1$, $1$ time to do exhaustive tuning, random-HP and SSL-HPT for one time series data, respectively. Let $p$ be the proportion of time series datasets used to train a neural network. It takes $O(pcN)$ to train a self-supervised learner using $pN$ time series datasets, let's ignore the time of training such light weight neural network for a moment, and it takes $O((1-p)N)$ to predict hyper-parameters for remaining $(1-p)N$ time series datasets. The total time cost is $O(pcN + (1-p)N)$. Similarly, we could get the cost of exhaustive tuning and random-HP are $O(cN)$ and $O(N)$. This suggests that, as $p$ goes to $1$, SSL-HPT turns out to be exhaustive tuning, and as $p$ goes to $0$, to be random-HP. The balance of these two approaches can be adjusted through $p$ based on users' needs!

\section{Applications of SSL-HPT}
\label{sec:apps}

In this section, we describe two approaches to utilize the SSL-HPT algorithm: (1). combine the SSL-MS and SSL-HPT algorithms to achieve large scale forecasting, and (2). leverage the SSL-HPT algorithm in ensemble models.

\subsection{Integrate SSL-MS with SSL-HPT}
\begin{figure*}
     \centering
     \includegraphics[width=0.9\textwidth]{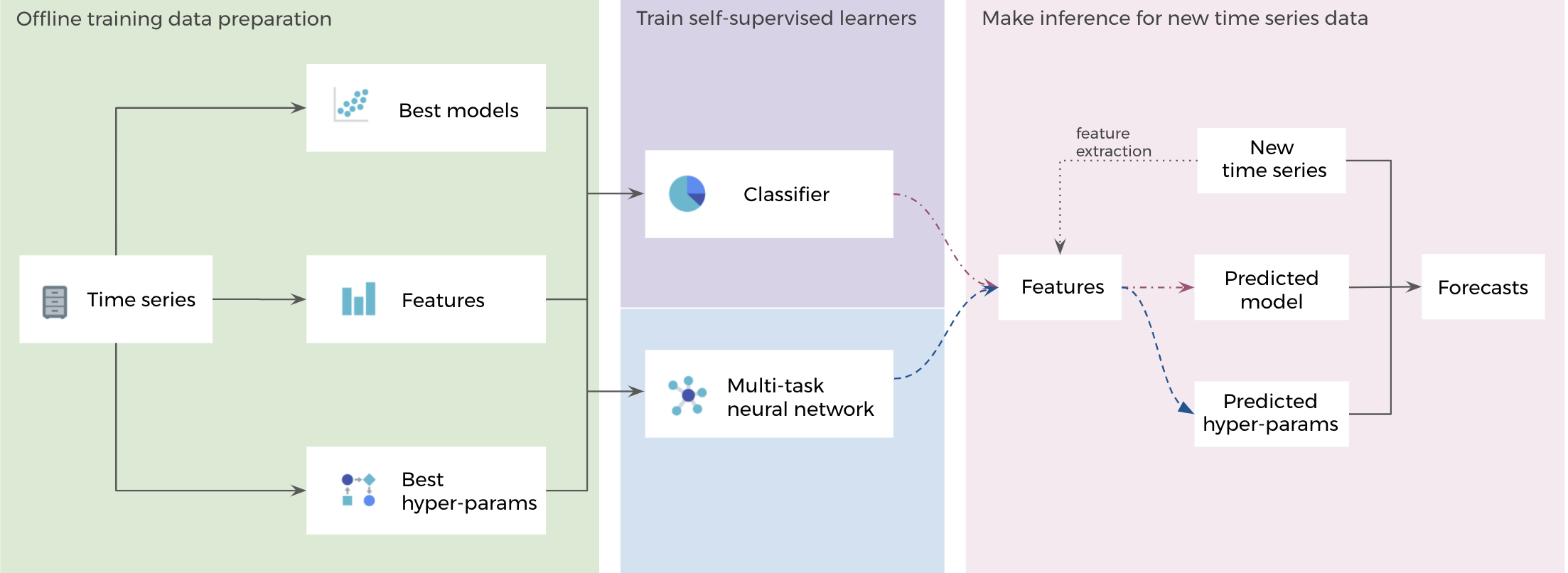}
      \caption{Workflow of SSL-MS with SSL-HPT}
      \label{fig:ssl-ms-ssl-hpt}
\end{figure*}

A straightforward application is to integrate SSL-MS with SSL-HPT. As Figure \ref{fig:ssl-ms-ssl-hpt} shows, after collecting offline training dataset, two SS-learners would be trained: one is for model selection, and the other for HPT. For a new time series data, we first predict a forecasting model to be used, and then hyper-parameters for this model. The computional time is constant, since both SSL-learners have been trained offline already. This makes it feasible to give fast and accurate forecasts at a large scale.

\subsection{Integrate ensemble models with SSL-HPT}
Ensemble is a learning process to leverage multiple learning algorithms to obtain a better predictive results than any of those multiple algorithms alone. It’s a practical way to reduce overfitting for complex models, because it combines a set of base learners together and then "smooth out" their predictions. 

We apply this process to time series forecasting, and leverage multiple time series models, such as ARIMA, SARIMA, Prophet \cite{taylor2018forecasting}, etc., to obtain final forecasts. Since the number of base learners is small, and a bad base learner influences the performance of the ensemble, a variant of the original ensemble algorithm \cite{hyndman2018forecasting} is proposed, called Median-Ensemble Model. Although this variant reduces negative effects from a bad learner, it’s still important to assure each base learner performs well, and therefore, a good parameters setting is needed for each model. 

\begin{figure*}
     \centering
     \includegraphics[width=0.9\textwidth]{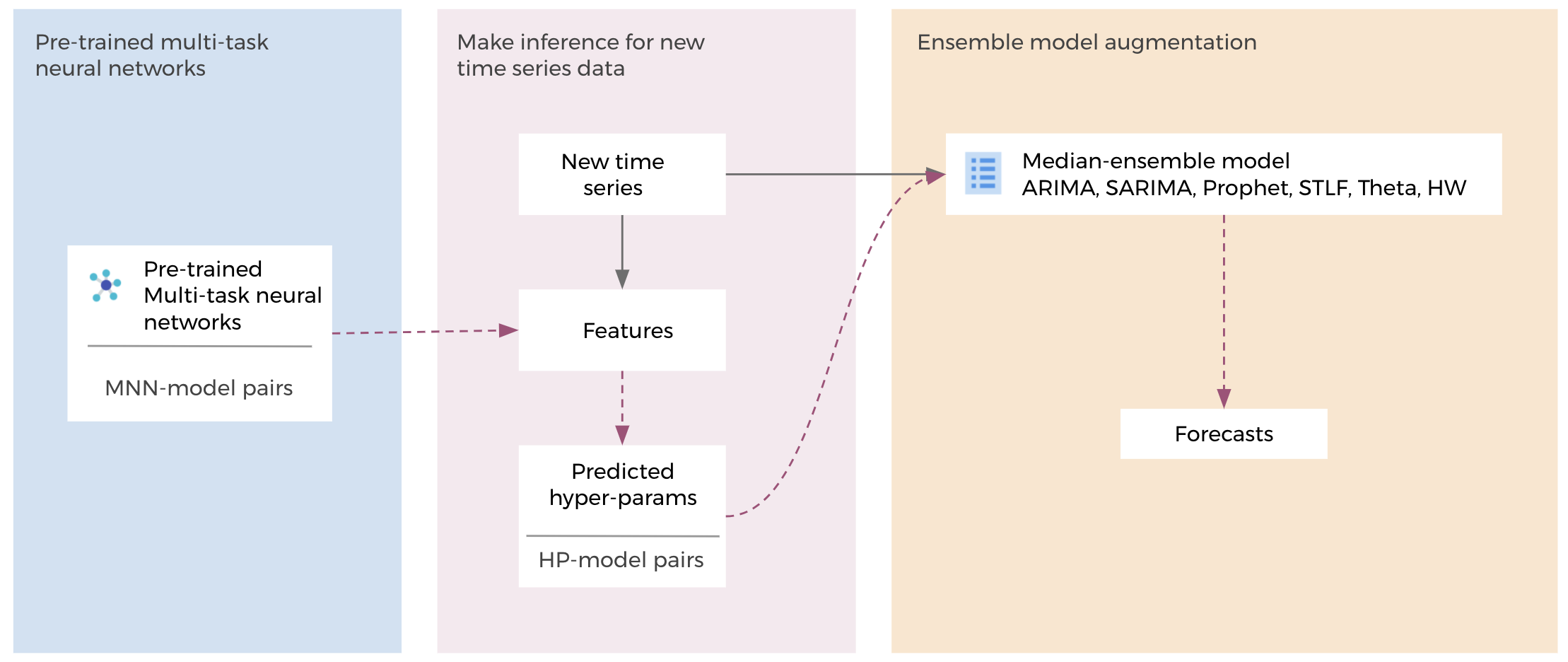}
      \caption{Workflow of ensemble model with SSL-HPT}
      \label{fig:ems-ssl-hpt}
\end{figure*}

Previously, parameters of each model in the ensemble are either given by the original HPT method, or randomly chosen from default search space. The first approach is accurate but very time consuming and expensive, while the second approach is faster but less accurate. Integrating ensemble models with SSL-HPT could balance these two approaches. Figure \ref{fig:ems-ssl-hpt} presents how to use predicted hyper-params from SSL-HPT instead of tuning all models in the ensemble.

%

\section{Experiment results}
\label{sec:expresults}

In this section, we present our experiment results with Facebook Infrastructure data, and M3-competition monthly data. Facebook data includes 2,852 time series from different Infrastructure domains, while M3-competition monthly data includes 1,428 time series from several areas, such as micro and macro finance, manufacturing, demographic, and so on. Currently, we are using 6 candidate forecasting models STLF, Theta \cite{assimakopoulos2000theta}, Prophet \cite{taylor2018forecasting}, ARIMA, SARIMA, and Holt-Winter’s model, which have widespread usage in the forecasting community.


In the offline-training data preparation step, we applied random search algorithm with 20 attempts, and then got the feature matrix $X$, 6 best-params matrices for 6 models: $Y_{STLF}$, $Y_{Theta}$, $Y_{Prophet}$, $Y_{ARIMA}$, $Y_{SARIMA}$, and $Y_{HW}$, and a best model matrix $Y$. Those meta-datasets are split into train and test sets. The training set is used for training 6 multi-task neural networks for ML-HPT and a classification model for SSL-MS, while test set is used for forecasts evaluation, where we also split into training and test sets.

\subsection{Results of forecasting on Facebook Infrastructure data}
\begin{figure}
     \centering
     \begin{subfigure}[b]{0.3\textwidth}
         \centering
         \includegraphics[width=\textwidth]{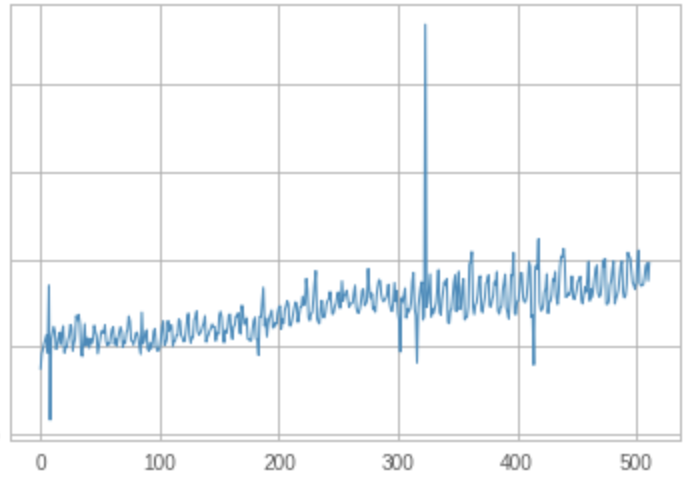}
     \end{subfigure}
     \hfill
     \begin{subfigure}[b]{0.3\textwidth}
         \centering
         \includegraphics[width=\textwidth]{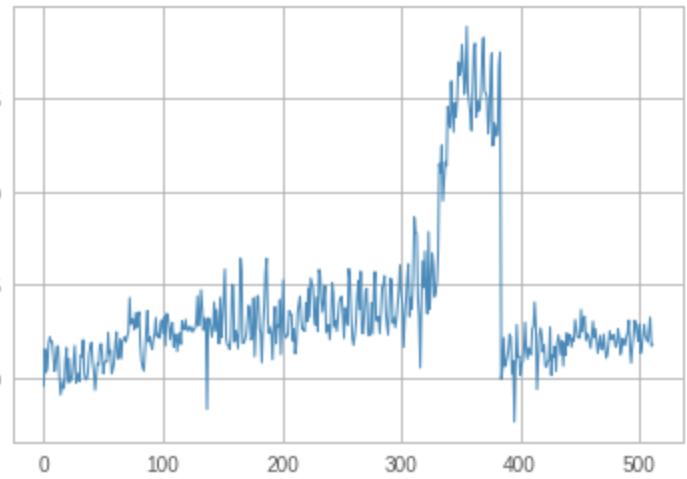}
     \end{subfigure}
     \hfill
     \begin{subfigure}[b]{0.3\textwidth}
         \centering
         \includegraphics[width=\textwidth]{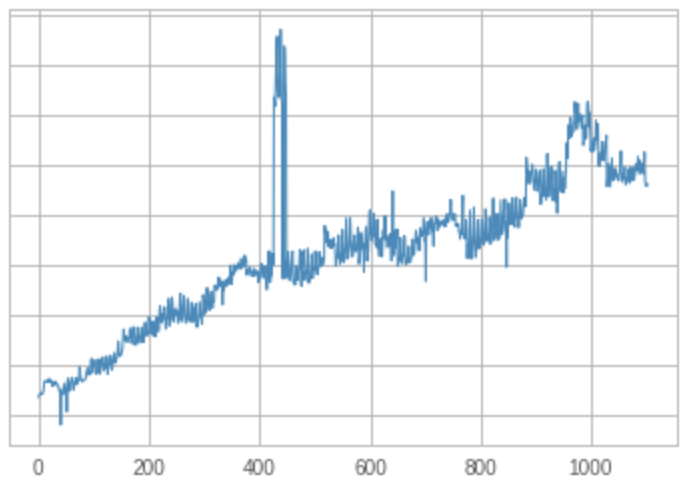}
     \end{subfigure}
        \caption{Time series plots examples on Facebook Infrastructure data}
        \label{fig:plt_infra}
\end{figure}

\begin{figure}
     \centering
     \includegraphics[width=0.4\textwidth]{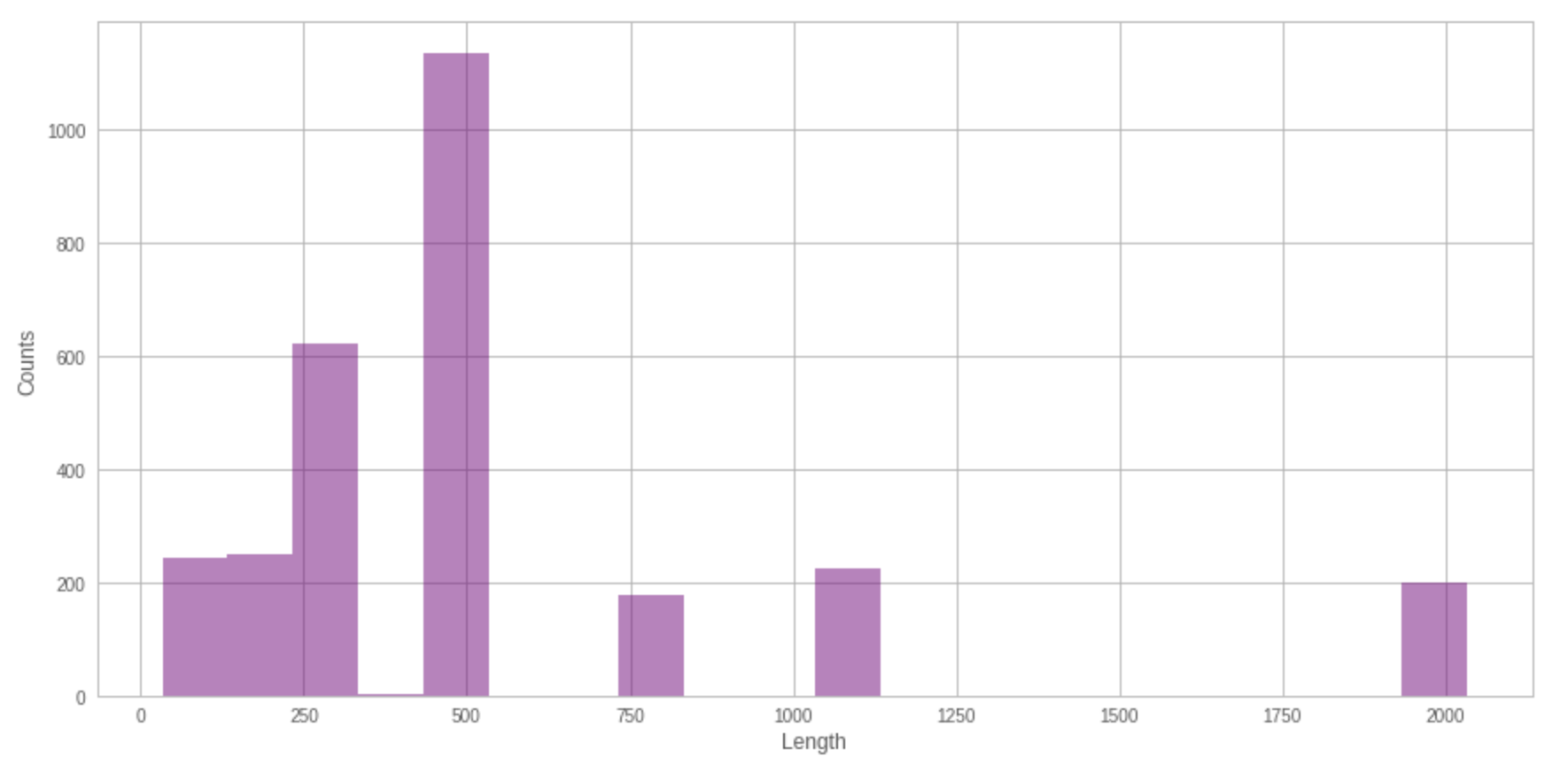}
      \caption{Distribution of time series lengths on Facebook Infrastructure data}
      \label{fig:ls_infra}
\end{figure}

Figure \ref{fig:plt_infra} shows three example plots of time series in Facebook data, and Figure \ref{fig:ls_infra} shows the distribution of their length. Typically, the appropriate forecasting model for a given time series depends on the number of observations. For example, shorter series tend to need simpler models such as a random walk, while for longer time series, we have enough information to be able to estimate a number of parameters for a complex model. Both Figure \ref{fig:plt_infra} and Figure \ref{fig:ls_infra} indicates that complicated models might be needed for those data.  

\begin{figure}
     \centering
     \begin{subfigure}[b]{0.45\textwidth}
         \centering
         \includegraphics[width=\textwidth]{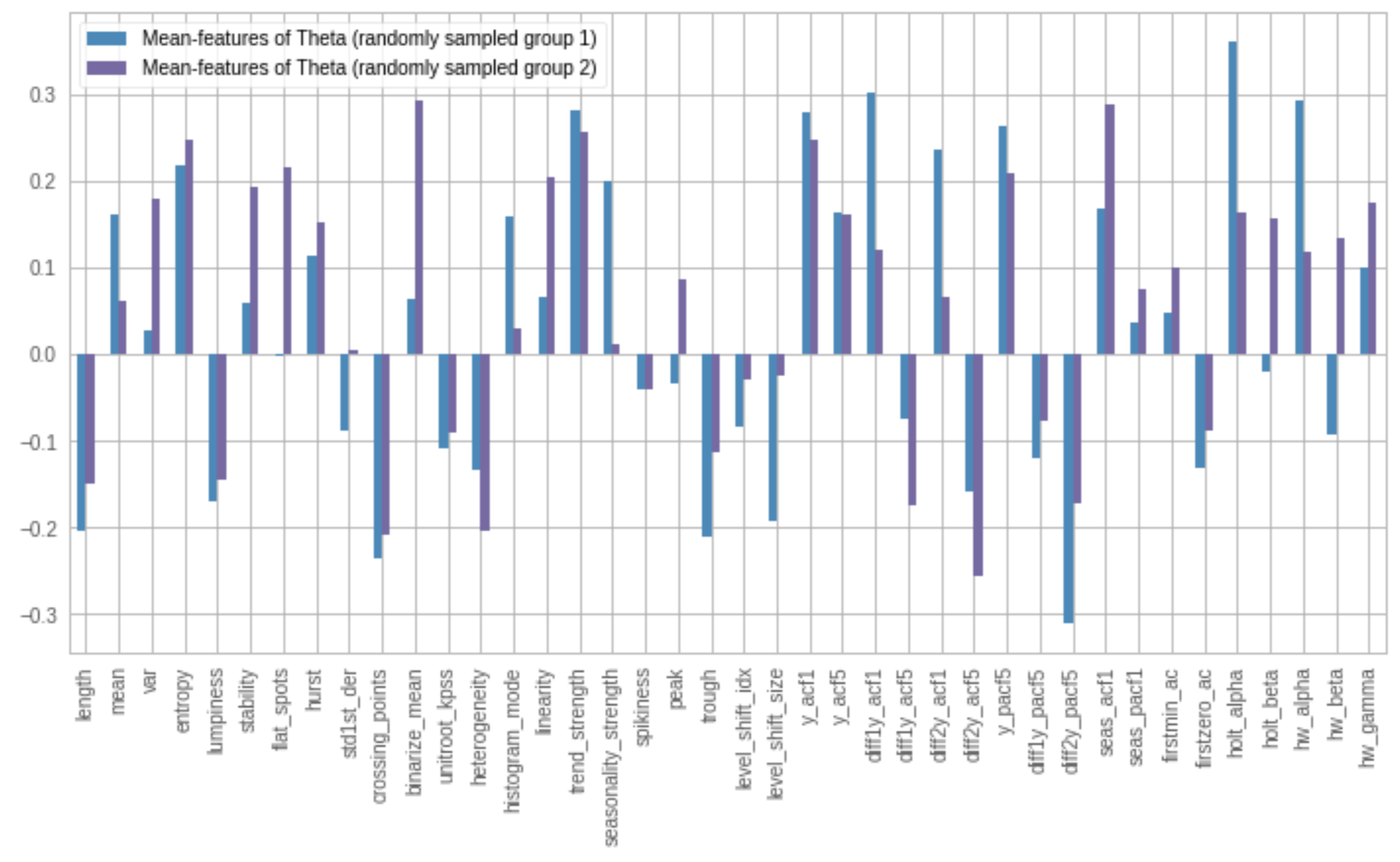}
     \end{subfigure}
     \hfill
     \begin{subfigure}[b]{0.45\textwidth}
         \centering
         \includegraphics[width=\textwidth]{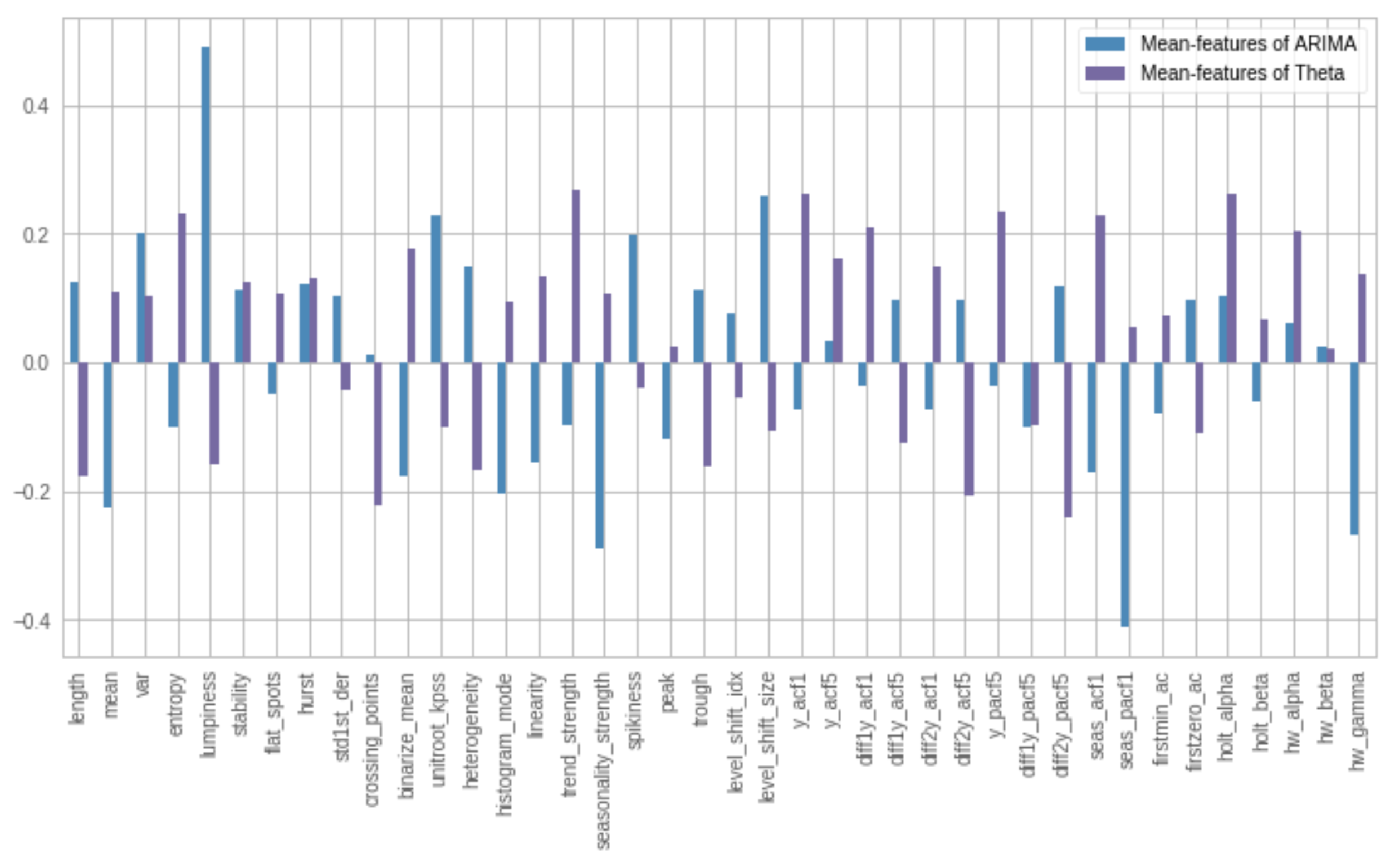}
     \end{subfigure}
        \caption{Features comparison of Facebook Infrastructure data with same and different selected models}
        \label{fig:fs_comp}
\end{figure}

Intuitively, time series datasets with similar features might be preferred by a specific model, and vice versa. We provide such feature comparison plots in Figure \ref{fig:fs_comp}. The bottom plot contains the bar plots of mean-feature-vector of time series datasets with label “ARIMA” and “Theta”, while the top one is the mean-feature-vector of randomly sampled half time series with label “Theta” and the other half with label “Theta”. It indicates that time series datasets with the same label have similar feature vectors and time series datasets with different labels have much more different feature vectors. This also provides interpretability of the model-features relationship, such as what are the features that a certain model favors? Moreover, it indicates the feasibility to train a SS-learner only based on features!

Compared with pre-selected model approach, we used almost the same computing time and resources, since the time of training a classifier could be ignored compared with hyper-parameter tuning. This is especially useful when we are making forecast periodically. Our aim is different from most classification problems in that we are not interested in accurately predicting the labels, but giving the best forecasts. It is possible that two models produce almost equally accurate forecasts, and therefore it is not important whether the classifier picks one model over the other. Therefore, in order to compare with the pre-selected models, we report the forecast's MAPE (Mean Absolute Percentage Error) on both training set and test set obtained from the SSL framework, and compare them with pre-selected models. The results are presented in Table \ref{table:mape1}.


\begin{table*}[]
\resizebox{0.9\textwidth}{!}{\begin{tabular}{c|c|c|c|c|c|c|c}
\hline
                                & SSL & Prophet                           & ARIMA     & STLF                              & SARIMA    & HW        & Theta     \\ \hline
Forecasts MAPE (train)     &\textbf{0.0871}	&0.1154	&0.1683	&0.1139	&0.1253	&0.1178	&0.1441 \\ \hline
MAPE Change \% (train) & \textbf{Baseline}      &-32.4811\%	&-93.1784\%	&-30.7371\%	&-43.8146\%	&-35.1966\%	&-65.4825\% \\ \hline
Forecasts MAPE (test)        & \textbf{0.1078} &0.1155	&0.1620	&0.1142	&0.1250	&0.1191	&0.1437   \\ \hline
MAPE Change \% (test)     &\textbf{Baseline}      &-7.1096\%	&-50.2112\%	&-5.9272\%	&-15.9224\%	&-10.4991\%	&-33.2728\% \\ \hline
\end{tabular}}
\caption{Performance comparison on Facebook Infrastructure data}
\label{table:mape1}
\end{table*}

Table \ref{table:mape1} shows that our algorithm performs best in terms of MAPE. We used random forest as our classification algorithm. MAPEs in Table \ref{table:mape1} are averaged values over 20 experiments for fair comparison. For the training set, it decreased MAPE by $30.74\%$ compared with the second best model STLF, and $18.35\%$ compared with Prophet. For the test set, it decreased MAPE by $5.93\%$ compared with STLF, and $7.11\%$ compared with Prophet. 

{\bf{Bounded empirical computational cost}}. Beyond the forecasting accuracy improvement, we actually don't need a large data set to train the SSL learner. In order to decide a feasible size of training set in the classification algorithm, we conducted extensive experiment and concluded that a training set with $size=1000$ time series would be sufficient to train a good SSL learner with Facebook Infrastructure data. Empirically, this means the computational cost is bounded for dataset with any numbers of time series.


{\bf{Consistentcy of model selection}}. We also checked the consistency of model selection results - if we choose different forecasting horizon, will the selected model be different? A consistent model for a specific time series data is preferred unless unexpected pattern appears in the data. For 200 time series with daily granularity, we predicted models at times $T$, $T + 0.5 \ \text{year}$, $T + 1 \ \text{year}$, $T + 1.5\  \text{years}$, $T + 2 \  \text{years}$. The results are presented in Table \ref{table:cons}, where $T = 1,303$ days. The first cell of this table means the model change rate at time $T$ and at time $T + 0.5 \ \text{year}$. For example, we predict models for 200 of them at Time T, half year later, we predict models for those time series again. We found $27\%$ of them have a different predicted model at two time points. Our results show that a predicted model is more likely to change after a longer period. From the diagonal cells, we could see that as $T$ grows larger, the half year change rate decreases, which means as the lenghth of the time series grows larger, the predicted model for this time series is becoming more stable.

\begin{table}[]
\resizebox{0.4\textwidth}{!}{\begin{tabular}{c|c|c|c|c}
\hline
          & T + y/2          & T + y            & T + 1.5 y        & T + 2 y          \\ \hline
T         & \textbf{27.00\%} & 34.50\%          & 33.50\%          & 42.50\%          \\ \hline
T + y/2   & NA               & \textbf{22.50\%} & 27.50\%          & 40.50\%          \\ \hline
T + y     & NA               & NA               & \textbf{17.00\%} & 37.00\%          \\ \hline
T + 1.5 y & NA               & NA               & NA               & \textbf{17.00\%} \\ \hline
\end{tabular}}
\caption{Consistency of model selection}
\label{table:cons}
\end{table}

Table \ref{table:all_comp1} shows results for all combinations of three model selection methods with three hyper-parameter tuning methods. Column “Avg-MAPE” and “Median-MAPE” represent averaged and median MAPEs over the test sets of all time series, respectively, column “$\#$ Fails” refers to the number of time series failed in model fitting due to the inappropriate hyper-parameters or model, and column “Runtime” denotes the theoretical runtime needed for each method. For example, the runtime of the first method is “1*20”, where “1” means costs for model selection, and “20” means costs for exhaustive HPT. We summarized our conclusions as follows.

\begin{itemize}
	\item[1. ]The ensemble model with exhausitive hyper-parameter tuning (HP) achieves the best performance overall (as expected), but our proposed SSL-MS with exhaustive HP has comparable forecasting performance with only 1/6 computational cost.
	\item[2. ]For any model selection approach, leveraging SSL-HPT always leads to forecasting performance improvement while having similar runtime.
	\item[3. ]What about SSL-MS + SSL-HPT? Surprisingly, this combination is our best choice (in terms of forecasting accuracy) if we need the lowest computational cost. 
\end{itemize}

\begin{table*}[]
\resizebox{\textwidth}{!}{\begin{tabular}{c|c|c|c|c|c|c|c|c}
\hline
					&            & Method                       & Avg-MAPE          & \% Avg-MAPE Change & Median-MAPE       & \% Median-MAPE Change & \# Fails   & Runtime          \\ \hline
  & 1          & Ensemble + HP &\textbf{0.111} &\textbf{-54.502\%}	&\textbf{0.096}	&\textbf{-38.407}\%	&0	& 6*20=120  \\ \cline{2-9}
			{Ensemble Model}			 & 2          & Ensemble + Random-HP &0.154	&-36.629\%	&0.130	&-16.517\%	&0	& 6*1=6 \\ \cline{2-9}
						 & 3          & Ensemble + SSL-HPT	&0.139	&-42.917\%	&0.116	&-25.379\%	&0	& 6*1=6  \\ \hline
						 
	 & 4          & Random model + HP    & 0.128	&-47.372\%	&0.104	&-32.976\%	&0        & 1*20=20          \\ \cline{2-9} 
			{Random Model}		& 5          & Random model + Random-HP     & 0.243          & \textbf{Baseline}    & 0.155         & \textbf{Baseline}       & 12  & 1*1=1            \\ \cline{2-9}
						& 6          & Random model + SSL-HPT       &0.178	&-26.889\%	&0.128	&-17.252\%	&3           & 1*1=1            \\ \hline
		& 7 		& SSL-MS + HP &\textbf{0.113}	&\textbf{-53.639}\%	&\textbf{0.090}	&\textbf{-42.174\%}	&0          &1*20=20 \\ \cline{2-9} 					
{SSL-MS}					& 8          & SSL-MS + Random-HP   &0.278	&14.536\%	&0.162	&4.020\%	&24         & 1*1=1            \\ \cline{2-9}
						& 9     & SSL-MS + SSL-HPT     &0.168	&-30.787\%	&0.124	&-20.401\%	&5          & 1*1=1            \\ \hline
\end{tabular}}
\caption{Overall comparison on Facebook Infrastructure data}
\label{table:all_comp1}
\end{table*}

\subsection{Results of forecasting on M3 data}
\begin{figure}
     \centering
     \begin{subfigure}[b]{0.3\textwidth}
         \centering
         \includegraphics[width=\textwidth]{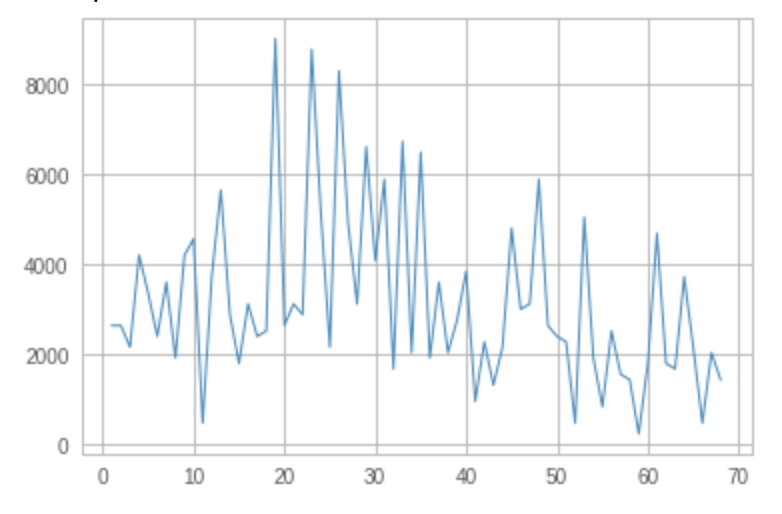}
     \end{subfigure}
     \hfill
     \begin{subfigure}[b]{0.3\textwidth}
         \centering
         \includegraphics[width=\textwidth]{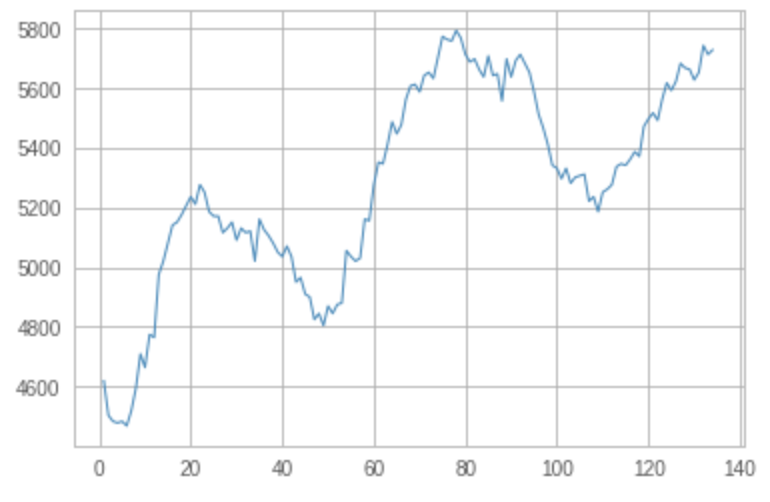}
     \end{subfigure}
     \hfill
     \begin{subfigure}[b]{0.3\textwidth}
         \centering
         \includegraphics[width=\textwidth]{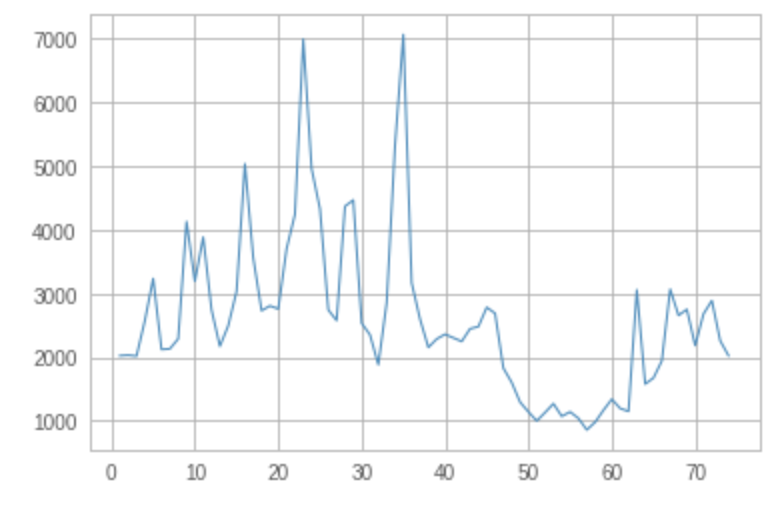}
     \end{subfigure}
        \caption{Time series plots examples on M3 data}
        \label{fig:plt2}
\end{figure}

\begin{figure}
     \centering
     \begin{subfigure}[b]{0.45\textwidth}
         \centering
         \includegraphics[width=\textwidth]{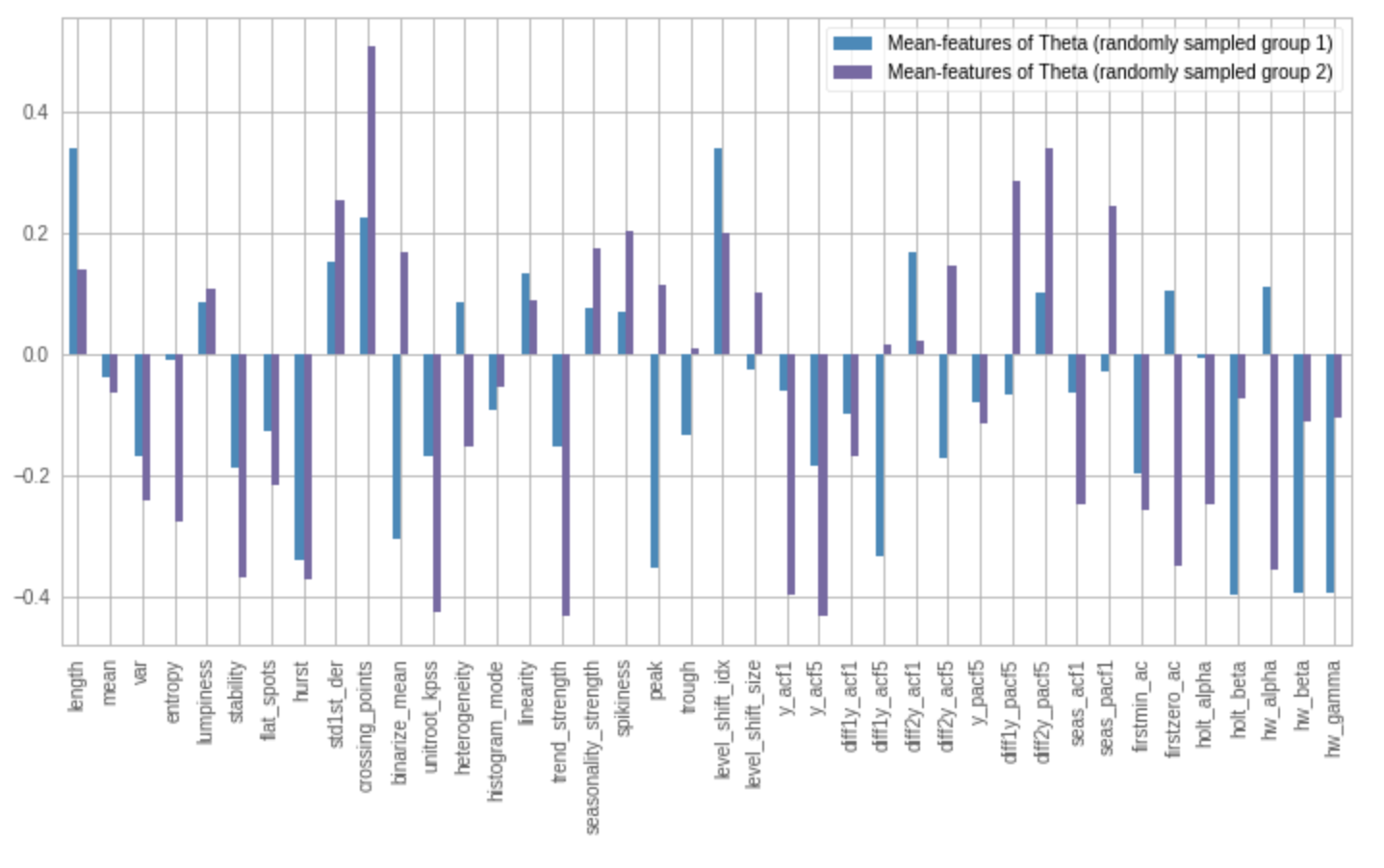}
     \end{subfigure}
     \hfill
     \begin{subfigure}[b]{0.45\textwidth}
         \centering
         \includegraphics[width=\textwidth]{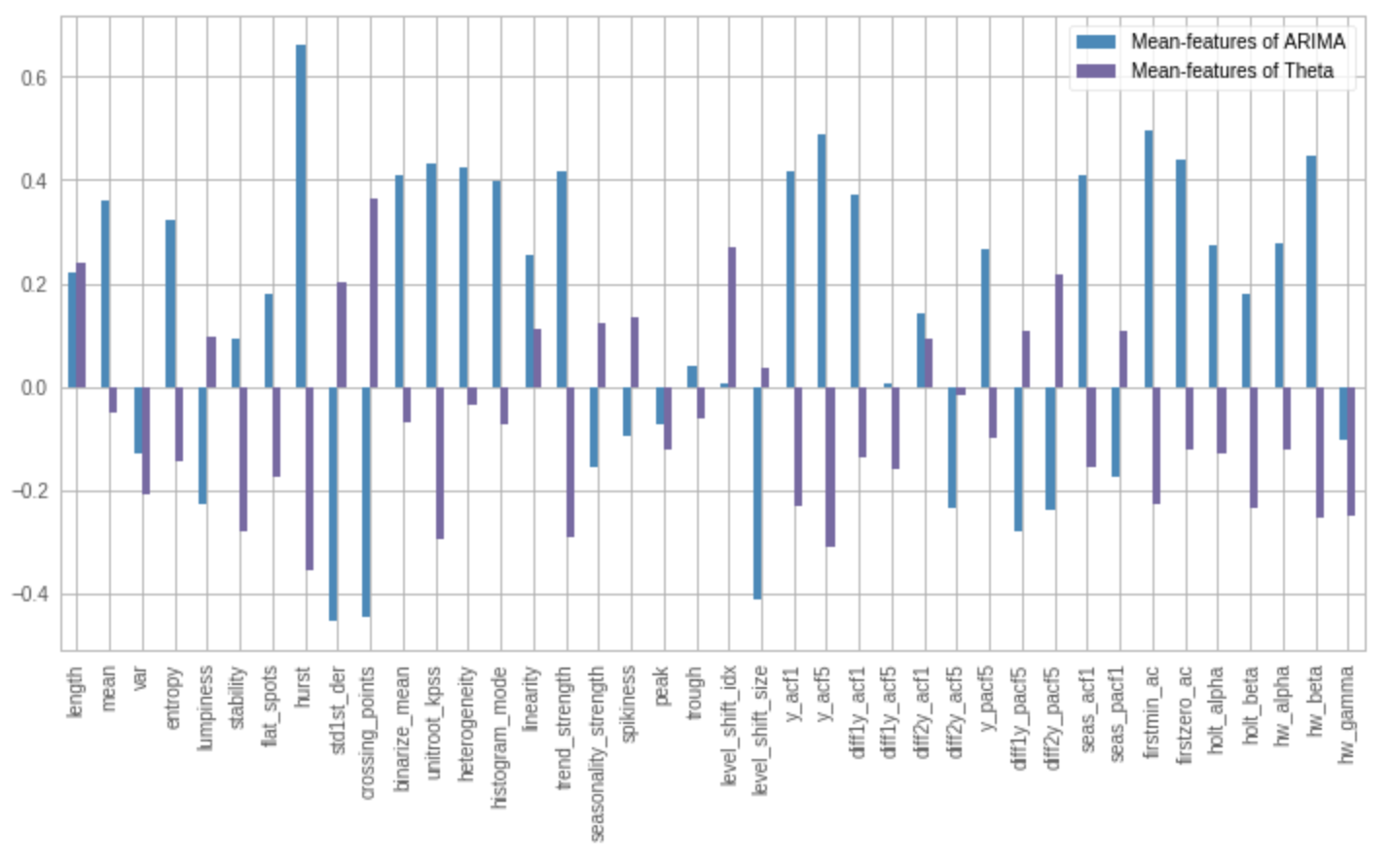}
     \end{subfigure}
        \caption{Features comparison plots on M3 data}
        \label{fig:fs_comp2}
\end{figure}

We observed similar results using M3-competition data. The lengths in this dataset are almost the same, with average length 110. Figure \ref{fig:plt2} shows time series plots examples of this dataset. Clearly, the patterns are much clearer compared with Facebook Infrastructure data. Figure \ref{fig:fs_comp2} shows features comparison plots, which are similar to our previous conclusion with Facebook Infrastructure data.

Table \ref{table:all_comp2} shows results with M3 data, with the same structure as Table \ref{table:all_comp1}. We can also obtain the similar conclusions as well. SSL-MS with exhaustive HP performs best across all experiments, even better than the ensemble with exhaustive HP approach. SSL-HP can always improve the forecasting performance compared with random-HP for all three model selection approaches. However, if we really need the approach with lowest computational cost, Random model with SSL-HPT and SSL-MS + SSL-HPT are comparably good, which demonstrates the effectiveness of the SSL-HPT method. The overall comparison table for M3-data shows the improvement is not as good as our results for Facebook Infrastructure data. This is expected because time series in M3 data are much shorter and simpler, and they're relatively easy to forecast.

\begin{table*}[]
\resizebox{\textwidth}{!}{\begin{tabular}{c|c|c|c|c|c|c|c|c}
\hline
							&            & Method                      	 & Avg-MAPE          & \% Avg-MAPE Change & Median-MAPE       & \% Median-MAPE Change & \# Fails   & Runtime          \\ \hline
  & 1          & Ensemble + HP 			&\textbf{0.098}	&\textbf{-42.727\%}		&\textbf{0.064}	&\textbf{-39.380}\%	&0	& 6*20=120  \\ \cline{2-9}
		{Ensemble Model}				 & 2          & Ensemble + Random-HP 	&0.131	&-23.252\%		&0.088	&-16.248\%	&0	& 6*1=6 \\ \cline{2-9}
						 & 3          & Ensemble + SSL-HPT		&0.123	&-27.969\%		&0.085	&-19.070\%	&0	& 6*1=6  \\ \hline
						 
	 & 4          & Random model + HP    		&0.106	&-38.043\%		&0.070	&-34.056\%	&0        & 1*20=20          \\ \cline{2-9} 
			{Random Model}			& 5          & Random model + Random-HP &0.170    & \textbf{Baseline}    & 0.105         & \textbf{Baseline}       & 11  & 1*1=1            \\ \cline{2-9}
						& 6          & Random model + SSL-HPT       &0.143	&-15.995\%		&0.097	&-7.945\%		&4           & 1*1=1            \\ \hline
						
		& 7 		& SSL-MS + HP 			&\textbf{0.091}	&\textbf{-46.770\%}	&\textbf{0.058}	&\textbf{-45.001\%}	&0          &1*20=20 \\ \cline{2-9} 					
		{SSL-MS}				& 8          & SSL-MS + Random-HP   	&0.168	&-1.582\%		&0.110	&4.758\%		&7         & 1*1=1            \\ \cline{2-9}
						& 9     & SSL-MS + SSL-HPT     		&0.145	&-14.726\%	&0.099	&-6.244\%		&4          & 1*1=1            \\ \hline
\end{tabular}}
\caption{Overall comparison on M3 monthly data}
\label{table:all_comp2}
\end{table*}

\section{Conclusion and Discussion}
\label{sec:conclusion}
In this paper, we proposed a \textbf{self-supervised learning (SSL)} framework for both model selection and hyper-parameter tuning, which provides accurate forecasts with lower computational time and resources. It contains three steps:

\begin{itemize}
\item[1. ]Offline training data preparation. Exhaustive tuning to obtain the best performed model for a given data, and hyper-parameters for each model and data combination. We empirically confirmed that this step only requires 1,000 time series data set.
\item [2. ]Offline self-supervised learners training. We train two learners - a classifier for model selection and a multi-tasks neural network for hyper-parameter tuning.
\item [3. ]Online inference (model selection and hyper-parameter tuning) for new time series data. 
\end{itemize}

Our experiment results show that integrating SSL-HPT with all model selection methods used almost same running time while improving accuracy compared with using random-HP. It reduces runtime compared with exhaustive HPT while maining a relatively good forecasting accuracy. By integrating SSL-MS with SSL-HPT, we achieves fast and scalable forecasts for a large collection of datases. 

\textbf{Extension on model recommendation.} Our proposed SSL framework can be easily extended to model recommendation systems, which means, in the model selection step, instead of giving a best forecasting model, we are giving a ranking of models. We borrow the idea from matrix factorization algorithm \cite{li_mf,lee1999learning,hoyer2004non,mnih2007probabilistic}, which is a very popular algorithm in recommendation system, and uses an embedding model. 

Suppose we have a set of users, and a set of items in a recommendation problem. Given the feedback matrix $A (m \times n)$, where $m$ is the number of users and $n$ is the number of items. The model is going to learn
\begin{itemize}
 \item[1. ]A user embedding matrix $U (m \times d)$, where row $i$ means the embedding for user $i$ 
 \item[2. ]An item embedding matrix $V (n \times d)$, where row $j$ means the embedding for item $j$
\end{itemize}
such that the product $UV^T$  is a good approximation of the matrix $A$. In our case, users are time series datasets, and items become candidate models. Matrix $A$ will be errors of forecasts, say MAPEs, where $a_{ij}$ is the $i$-th time series’ forecasts MAPE using model $j$ in $A$. Matrix $U$ will be time series features matrix, where $u_i$ is the features vector of the $i$-th time series. Our goal is to estimate matrix $V$.
Once a new time series data comes, we calculate the feature vector $u^*$, then the product of $u^* V^T$ will return an MAPE vector and thus we can provide a model ranking.

%

%
\textbf{Integrate SSL-HPT with Bayesian Optimal Search. }In contrast to Random or Grid Search, Bayesian Optimal Search (BOS) \cite{balandat2020botorch,feurer2018scalable} is able to keep track of past evaluation results and choose the next hyper-parameters in an informed manner, and thus be more efficient. Given an initial point, BOS iteratively searches for next best point until max iterations is reached. Although a global optimization technique would be able to find out global optima irrespective of initial values, it might take more iteration steps if using a “bad” initial values of parameters. Based on our early experiments, our proposed SSL-HPT algorithm provides good initial values for BOS. Given the same iteration times, the combination of BOS and SSL-HPT (SSL-HPT-BOS) achieved better performance than BOS alone. This paves a path to generalize our SSL framework to generic hyper-parameter tuning for any statistical and machine learning models in the future.

\section{Acknowledgments}

\begin{acks}
We would like to thank Alessandro Panella and Dario Benavides for the valuable discussion and feedback.
\end{acks}

\bibliographystyle{ACM-Reference-Format}
\bibliography{ssl}

\appendix

\section{List of time series features}
\label{appendixA}
\begin{itemize}
 \item Length: length of time series data. 
 \item  Mean/Variance (2): mean/variance of time series.
 \item  Spectral entropy: Shannon entropy of spectral density function. High value indicates noisy and unpredictable time series.
 \item  Lumpiness: variance of variances within non-overlapping windows
 \item  Stability: variance of means within non-overlapping
 \item  Trend/Seasonal strength (2): measures of trend and seasonality of a time series based on an STL decomposition. Trend strength is the variance explained by STL trend term, and seasonal strength is the variance explained by STL seasonality terms.

 \item  Spikiness: variance of leave-one-out variances of STL remainder after an STL decomposition.
 
 \item  Peak/Trough (2): location of peak/trough in the seasonal component after an STL decomposition.

 \item  Flat spots: presence of flat segments. It’s computed by dividing the sample space of a time series into nbins equal-sized intervals, and computing the maximum run length within any single interval.
 
 \item  Level shift based features (2): These two features compute features of a time series based on sliding (overlapping) windows. 
 \item  Hurst Exponent is used as a measure of long-term memory of time series. It relates to the autocorrelations of the time series, and the rate at which these decrease as the lag between pairs of values increases. 
 \item  ACF based features (7): summarizes the strength of a relationship between an observation in a time series with observations at prior time steps. We compute ACs of the series, the differenced series, and the twice-differenced series, and then provide a vector comprising the first AC in each case, the sum of squares of the first 5 ACs in each case, and the AC at the first seasonal lag.
 \item  PACF based features (4):summarizes the strength of a relationship between an observation in a time series with observations at prior time steps with the relationships of intervening observations (indirect correlation) removed. We compute PACs of the series, the differenced series, and the second-order differenced series, and then provide a vector comprising the sum of squares of the first 5 PACs in each case, and the PAC at the first seasonal lag.
 \item  First min AC: the time of first minimum in the autocorrelation function.
 \item  First zero AC: the time of first zero crossing the autocorrelation function.
 \item  Linearity: R square from a fitted linear regression.
 \item  Standard deviation of the first derivative of the time series
 \item  Crossing Points: the number of times a time series crosses the median line.
 \item  Binarize mean: converts time series into a binarized version: time series values above its mean are given 1, and those below the mean are 0, and then returns the average value of the binarized vector.
 \item  ARCH statistic: Lagrange multiplier test statistic from Engle’s Test for Autoregressive Conditional Heteroscedasticity (ARCH). It measures the heterogeneity of the time series.
 \item  Histogram mode: measures the mode of the data vector using histograms with a given number of bins.
 \item  KPSS unit root statistic: a test statistic based on KPSS test, which is to test a null hypothesis that an observable time series is stationary around a deterministic trend. Linear trend and lag one are used here.
 \item  Holt Parameters (2): estimates the smoothing parameter for the level-alpha and the smoothing parameter for the trend-beta of Holt’s linear trend method.
 \item  Holt-Winter’s Parameters (3): estimates the smoothing parameter for the level-alpha, trend-beta of HW’s linear trend, and additive seasonal trend-gamma.
  \end{itemize}

\end{document}